\documentclass[authoryear,preprint,review,12pt]{elsarticle}

\usepackage{amssymb}
\usepackage{amsmath}

\usepackage{tikz,graphics,color,epsf,caption,subcaption}
 \usepackage{float}
\usepackage{graphicx}
\usepackage{url}
\usepackage{multirow}
\usepackage{booktabs}

\journal{Engineering Application of AI}

\begin{document}
\begin{frontmatter}



\title{Multi-Agent Reinforcement Learning for Heterogeneous Satellite Cluster Resources Optimization} 


\author[label1]{Mohamad A. Hady} 
\ead{mohamad.had@mymail.unisa.edu.au}

\author[label2]{Siyi Hu}
\ead{siyi.hu@curtin.edu.au}

\author[label1]{Mahardhika Pratama}
\ead{dhika.pratama@unisa.edu.au}

\author[label1]{Zehong Cao}
\ead{jimmy.cao@unisa.edu.au}

\author[label1,label3]{Ryszard Kowalczyk}
\ead{ryszard.kowalczyk@unisa.edu.au}

\affiliation[label1]{organization={STEM, University of South Australia},
            addressline={Mawson Lakes Blvd}, 
            city={Mawson Lakes},
            postcode={5095}, 
            state={SA},
            country={Australia}}
\affiliation[label2]{organization={School of Electrical Engineering, Computing and Mathematical Sciences (EECMS), Curtin University},
            addressline={Kent St}, 
            city={Bentley},
            postcode={6102}, 
            state={WA},
            country={Australia}}
\affiliation[label3]{organization={Systems Research Institute,Polish Academy of Sciences},
            addressline={}, 
            city={Warsaw},
            postcode={}, 
            state={},
            country={Poland}}

\begin{abstract}
This work investigates resource optimization in heterogeneous satellite clusters performing autonomous Earth Observation (EO) missions using Reinforcement Learning (RL). In the proposed setting, two optical satellites and one Synthetic Aperture Radar (SAR) satellite operate cooperatively in low Earth orbit to capture ground targets and manage their limited onboard resources efficiently. Traditional optimization methods struggle to handle the real-time, uncertain, and decentralized nature of EO operations, motivating the use of RL and Multi-Agent Reinforcement Learning (MARL) for adaptive decision-making. This study systematically formulates the optimization problem from single-satellite to multi-satellite scenarios, addressing key challenges including energy and memory constraints, partial observability, and agent heterogeneity arising from diverse payload capabilities. Using a near-realistic simulation environment built on the Basilisk and BSK-RL frameworks, we evaluate the performance and stability of state-of-the-art MARL algorithms such as MAPPO, HAPPO, and HATRPO. Results show that MARL enables effective coordination across heterogeneous satellites, balancing imaging performance and resource utilization while mitigating non-stationarity and inter-agent reward coupling. The findings provide practical insights into scalable, autonomous satellite operations and contribute a foundation for future research on intelligent EO mission planning under heterogeneous and dynamic conditions.

\end{abstract}



\begin{keyword}


Reinforcement Learning \sep Multi-agent Reinforcement Learning \sep Heterogeneous Agent \sep Heterogeneous Satellite Systems \sep Resources Optimization \sep Earth Observation mission.

\end{keyword}

\end{frontmatter}



\section{Introduction}
The rapid expansion of Low Earth Orbit (LEO) satellite constellations has greatly enhanced Earth Observation (EO) capabilities, supporting applications such as climate monitoring, disaster management, agricultural assessment, and urban planning. However, coordinating and managing multiple satellites autonomously remains a fundamental challenge due to the dynamic, uncertain, and resource-constrained nature of space environments \citep{wang2020agile, chen2019mixedILP, stephenson2023optimal, pan2023dense}. Unlike pre-planned mission operations, autonomous EO missions require each satellite to make real-time decisions under partial observability, fluctuating environmental conditions, and strict resource limitations, while maintaining coordinated behaviour across the entire constellation \citep{li2024mission, yang2024objective}. These challenges stem from several interacting factors: uncertainty in observation conditions (e.g., variations in solar exposure and cloud coverage affecting energy availability and imaging quality), limited onboard resources such as power, data storage, and attitude control momentum, and the intrinsic non-stationarity of multi-agent settings, where each satellite’s actions continuously alter the shared environment and other agents’ states \citep{araguz2018applying, yao2019task}.

Beyond these general coordination difficulties, EO missions increasingly employ heterogeneous satellite clusters that combine different payload types such as Synthetic Aperture Radar (SAR) and optical sensors \citep{cohen2017novasar, dong2024optisar, alzubairi2024spacecraft}. SAR sensors can operate under cloud coverage and at night, while optical sensors offer high-resolution imaging under clear conditions. This complementary capability substantially improves temporal coverage and observation reliability but introduces new coordination and control complexities. Differences in field of view, imaging geometry, and operational constraints lead to divergent observation and action spaces across satellites, creating heterogeneity at both the physical and decision-making levels. Managing such a mixed constellation thus requires not only efficient resource scheduling but also adaptive policies that account for diverse capabilities and dynamics.

\begin{figure}[ht!]
    \centering
    \includegraphics[width=\textwidth]{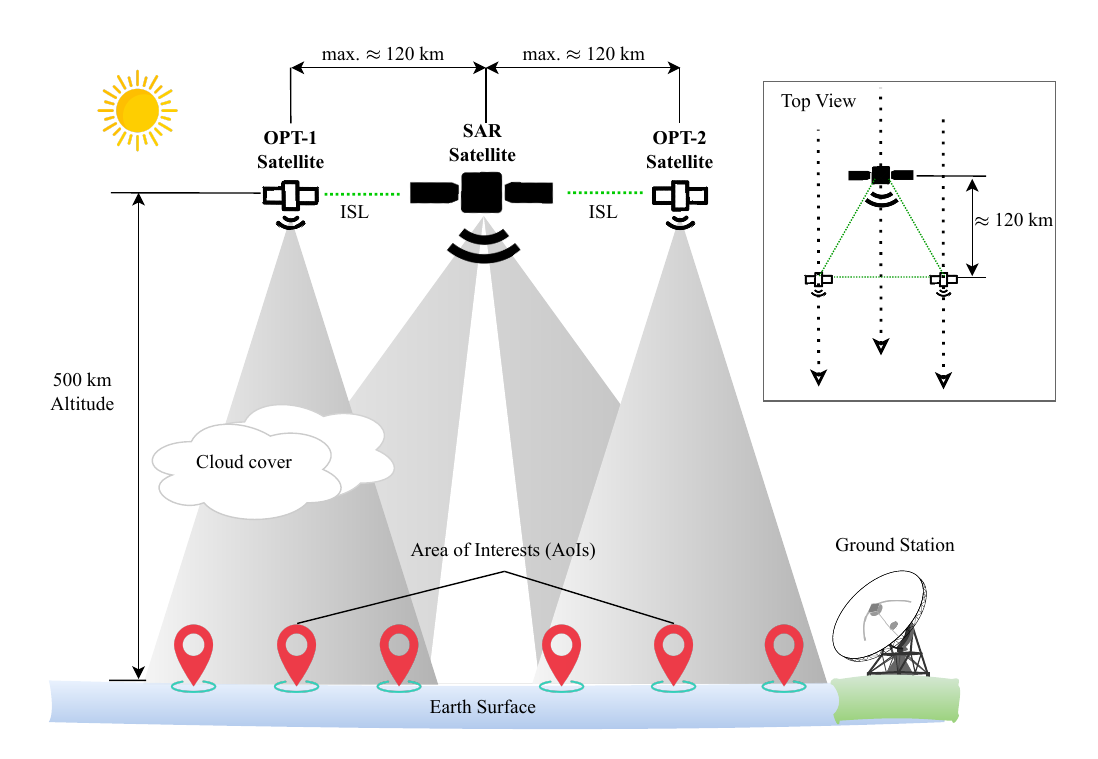}
    \caption{Heterogeneous satellite cluster EO Mission with SAR and Optical (OPT) satellite. The satellites are deployed in three different narrow orbits forming a cooperative formation. SAR sensor will be used to cover regions with higher cloud coverage that is the burden of Optical sensor. Therefore, coordination in resource optimization across heterogenous satellite is important in this case and raise new challenge in the autonomous operation. Our scenario is implemented in BSK-RL and the Basilisk Simulator \citep{stephenson2024bsk}. It is a realistic satellite system simulator with 3D visualization (Vizard).}
    \label{fig:framework}
\end{figure}

Traditional optimization techniques, such as mixed-integer linear programming (MILP), have been explored for constellation scheduling and resource allocation \citep{kim2024optimal}. While effective in static conditions, these approaches rely on predefined models and are limited in their ability to adapt to time-varying environmental and operational uncertainties. Reinforcement Learning (RL) provides a promising alternative by enabling agents to learn adaptive strategies through interaction with the environment, allowing satellites to optimize imaging, energy usage, and data management under uncertainty \citep{herrmann2024single, stephenson2024bsk}. However, as EO missions scale from single satellites to cooperative constellations, the decision-making problem expands from isolated optimization to distributed coordination among multiple agents — a setting more naturally modeled by Multi-Agent Reinforcement Learning (MARL) \citep{tang2024dynamic}.

MARL offers a framework for decentralized decision-making, where each satellite acts as an autonomous agent learning to coordinate with others through shared experience. Existing MARL applications in satellite operations \citep{herrmann2023reinforcement, stephenson2024reinforcement} have shown encouraging results but often rely on fully centralized training or continuous communication assumptions, which are impractical for real-world missions with limited inter-satellite links. Several learning architectures have emerged to mitigate this, including Fully Centralized (CTCE), Fully Decentralized (DTDE), and Centralized Training with Decentralized Execution (CTDE) paradigms \citep{ning2024survey, hady2025multi}. Among them, CTDE achieves a practical balance by enabling satellites to learn coordinated policies during centralized training while operating independently at execution time. Recent on-policy methods such as Multi-Agent Proximal Policy Optimization (MAPPO) have demonstrated stable performance across diverse MARL environments \citep{yu2022surprising}.

Despite these advances, current MARL frameworks largely assume homogeneous agents, where all satellites share identical dynamics and observation–action structures. This assumption limits applicability to realistic EO missions, where sensor diversity and operational asymmetry fundamentally change the learning problem. Recent progress in heterogeneous-agent MARL, such as Heterogeneous-Agent PPO (HAPPO) \citep{kuba2022trust, zhong2024heterogeneous}, introduces separate value estimation and policy update mechanisms to address agent heterogeneity. Yet, their performance and adaptability in physically grounded, resource-constrained satellite systems remain unexplored. The interaction between heterogeneous payloads, limited energy, and asynchronous decision-making introduces new learning instabilities and coordination trade-offs not captured in prior work.

In this paper, we present a comprehensive study of heterogeneous satellite resource optimization in autonomous EO clusters using reinforcement learning. Building upon realistic satellite dynamics simulated in BSK-RL and the Basilisk platform \citep{stephenson2024bsk}, we develop and evaluate MARL frameworks that explicitly account for agent heterogeneity. The general scenario of heterogeneous cluster is illustrated in Fig.~\ref{fig:framework}.

Our key contributions are as follows:
\begin{itemize}
\item \textbf{A structured modeling framework for heterogeneous satellite cluster resource optimization.} We formulate the problem to capture satellite-specific operational differences, including varying state transitions, sensor geometries, and resource constraints. To our knowledge, this is the first study to model heterogeneity explicitly in a realistic satellite cluster as a multi-agent learning problem.
\item \textbf{An empirical investigation of learning dynamics and coordination challenges} across single-satellite, homogeneous, and heterogeneous configurations. We analyze how factors such as resource coupling, sensor diversity, and inter-agent dependencies influence convergence, stability, and performance.

\item \textbf{A comparative evaluation of state-of-the-art RL and MARL algorithms} — including PPO, MAPPO, HAPPO, and HATRPO — applied to the heterogeneous cluster setting. The results highlight the strengths and limitations of existing methods and provide insights into how heterogeneity affects scalability and adaptability in multi-satellite coordination. All code and demonstration videos are publicly available.\footnotemark{https://anonymous.4open.science/r/hetclusterF00A/}
\end{itemize}

The rest of this paper is structured as follows: Section II presents the heterogeneous satellite resources optimization. Section III describes the framework algorithm design  employed to solve the problem starting from a single-satellite to satellite cluster scenarios. Section IV discusses our experimental evaluation and results. Finally, Section V concludes the paper and discusses future directions.

\section{Heterogeneous Satellite Cluster Resources Optimisation}
Real-time resources optimization can be seen as an optimization or search of a policy for sequential decision making process. Instead of optimizing every decision in each time-step, we can design an optimal policy along with the sequential decision making processes or well-known as Markov Decision Process (MDP). This model extension to the multi-agent setting is mainly discussed in this section. Firstly, the optimization objective function is designed with constraints and then followed by the challenges. The heterogeneity of the satellite cluster aspect is also formulated in the sense of heterogeneous agents.

\subsection{Satellite Resources Optimization}
The total objective consists of two main components: (i) data acquisition and 
(ii) resource utilisation. The goal is to maximize mission value while ensuring 
efficient use of on-board resources. The overall objective is defined as:
\begin{equation}
\max J(Q, F, x, t) = Q + \sum_{t=0}^{T} \sum_{j=1}^{M} \sum_{k=1}^{N} F_{j,k}\big(x_{j,k,t}\big),
\end{equation}
where, $Q$ denotes a constant base reward, $T$ represents the length of the planning horizon, $M$ is the number of satellites (or agents), $N$ is the number of tasks (or targets), $x_{j,k,t}$ is the binary decision variable indicating whether satellite $j$ is assigned to task $k$ at time $t$, $F_{j,k}(x_{j,k,t})$ denotes the reward contribution associated with assigning task $k$ to satellite $j$ at time $t$.

In this objective, the goal is to maximize the total accumulated reward over all agents and tasks within the time horizon, in addition to the base reward $Q$. The triple summation term captures the spatio-temporal reward contributions resulting from assignment decisions across all $(j,k,t)$ combinations.

\subsubsection{Objective Function Design}
\paragraph{1) Data Acquisition (High-Priority AoIs)}
The first objective is to prioritise the most valuable targets. This term is expressed as
\begin{equation}
\max Q = \max \sum_{i=1}^{T} q_i,
\end{equation}
where $q_i$ denotes the reward for acquiring data from a high-priority AoI. Increasing $Q$ 
corresponds to prioritizing the most important targets.

\paragraph{2) Resource Utilisation}
The resource-related component $F_{j,k}$ is composed of three sub-objectives:

\textit{$\bullet$ minimize Power Usage ($F_{1,k}$):}
    \begin{equation}
    \min F_{1,k} = \min \sum_{t=0}^{T} \sum_{i=1}^{N} f_{1,k}\big(x_{1,i,t}\big),
    \end{equation}
    promoting efficient energy consumption.

\textit{$\bullet$ maximize Data Downlinking ($F_{2,k}$):}
    \begin{equation}
    \max F_{2,k} = \max \sum_{t=0}^{T} \sum_{i=1}^{N} f_{2,k}\big(x_{2,i,t}\big),
    \end{equation}
    encouraging more data to be successfully transmitted to the ground.

\textit{$\bullet$ maximize Payload Usage ($F_{3,k}$):}
    \begin{equation}
    \max F_{3,k} = \max \sum_{t=0}^{T} \sum_{i=1}^{N} f_{3,k}\big(x_{3,i,t}\big).
    \end{equation}
    It is designed to ensure the correct selection of sensing payloads (e.g., SAR for cloudy conditions and optical for clear scenes) and to provide useful data for the user.

\subsubsection{Satellite Resources Constraints}
A single satellite has two limited resources that are considered as \textit{constraints} in our study: battery level ($b_t \in [B_{min},B_{max}]$) and data storage capacity ($d_t \in [D_{min},D_{max}]$) at any time step ($t$). At each time step, the satellite consumes electrical power, denoted as $c_{b,i}$, and stores data, represented as $c_{d,i}$. The battery is rechargeable via a solar panel. To maximize battery charging, the satellite must adjust its attitude toward the sun, which may conflict with its target imaging orientation.
Another constraint arises from attitude control, specifically the speed of the Reaction Wheels (RWs), denoted as $\hat{\Omega} \in [-\Omega_{max},\Omega_{max}]$. These wheels serve as the primary actuators for satellite attitude adjustments along the three axes ($x,y,z$). To prevent exceeding the maximum speed threshold, the satellite must periodically desaturate the wheels.
The limited resource constraints are expressed as:
\begin{equation}
    \sum^{\infty}_{t=0}{c_{b,t}}\leq b_t, \: B_{min}\leq b_t \leq B_{max} \quad
     \text{and} \quad \sum^{\infty}_{t=0}{c_{d,t}}\leq d_t, \: D_{min}\leq d_t \leq D_{max}.
\end{equation}

These constraints are incorporated into the model as \textit{failure} (Eq. \ref{eq:failure}) triggers in the reward function, resulting in penalties or negative rewards as feedback from the environment. Some other constraints, such as the communication baud rate, have a relevant impact on the system performance. However, in this work, it is assumed as fixed in time as the transmitter specification.

\subsection{Satellite Cluster Modelling}
A single EO satellite functions as an agent, making decisions at discrete time steps based on its current state. The simulation defines four possible actions: \textit{1) Capturing the $i$-th image target}, where the satellite must orient its optical imaging sensor toward a selected target-$i$ among the available targets on Earth and store it in the on-board memory; \textit{2) Downlinking}, where the satellite transmits the collected EO image data whenever it has access to a ground station; \textit{3) Charging}, which involves reorienting the satellite toward the sun to maximize solar energy absorption and recharge its battery; \textit{4) Desaturating}, which ensures that the Reaction Wheels (RWs), the primary actuators for attitude control, operate within safe speed limits. If the RW speed approaches saturation, the satellite must execute a desaturation maneuver to maintain stable attitude control and prevent uncontrollable drift.

To formally represent the autonomous satellite decision-making process as a POMDP, we define the problem as a tuple:  
\begin{equation}
    \mathcal{G}=\langle\mathcal{S,A,O,T},r,\mathcal{Z,\gamma}\rangle,
\end{equation}
where $\mathcal{S}$ represents a finite set of states that defines the true underlying condition of the satellite in space, which is not fully observable. The entire state representation is encapsulated in the Basilisk simulator, including all relevant physical parameters necessary for EO satellite operations. $\mathcal{A}$ is a finite set of available actions, including capturing, downlinking, charging, and desaturating. $\mathcal{O}$ is the finite set of observations the agent can receive, such as battery level, data storage availability, reaction wheel speed, target opportunity window, ground station access window, eclipse status, and time.
The transition probability function $\mathcal{T}(s_t,a_t,s_{t+1})=P(s_{t+1}|s_t,a_t)$ defines the probability of transitioning to state $s_{t+1}$ from $s_t$ after executing action $a_t$. The reward function $r$ determines the immediate reward for taking an action in a given state. The observation probability function $\mathcal{Z}(o_t,s_{t+1},a_t)=P(o_t|s_{t+1},a_t)$ defines the probability of observing $o_t$ given that the system is in state $s_{t+1}$ after taking action $a_t$. Finally, the discount factor $\mathcal{\gamma} \in [0,1]$ determines the importance of future rewards.

The use of multiple satellites in a cluster formation enhances the capabilities of EO missions. These satellites can be viewed as a multi-agent system, particularly when they collaborate to achieve common mission objectives. A multi-agent system can be modelled as a \textit{Multi-Agent Markov Decision Process (MA-MDP)} or a \textit{Stochastic Game} \citep{hu2003nash}, which is defined by the tuple:
\begin{equation}
    \mathcal{M}=\langle \mathcal{N},\mathcal{S},\{\mathcal{A}_i\}_{i \in \mathcal{N}},\{r_i\}_{i \in \mathcal{N}},\mathcal{T},\gamma \rangle,
\end{equation}
where $\mathcal{N}$ represents the set of $n$ agents, $\mathcal{S}$ is the state space of the environment, and $\mathcal{A}_i$ is the action space of agent-$i$, with the joint action space defined as $\mathcal{A}=\mathcal{A}_1 \times \mathcal{A}_2 \times \dots \times \mathcal{A}_n$. The state transition probability function $\mathcal{T}: \mathcal{S} \times \mathcal{A} \times \mathcal{S} \rightarrow [0,1]$ defines the likelihood of transitioning between states given a joint action. The reward function $r_i: S \times A \rightarrow \mathbb{R}$ determines the individual reward received by each agent-$i$. 

In stochastic games, the strategic relationships between agents can be categorized into three distinct types: cooperative, competitive and mixed \citep{zhang2021multi}. In \textit{cooperative} type, all agents work toward a shared objective and optimize a global reward function. This study primarily focuses on cooperative settings, where satellites collaborate to capture unique EO images while sharing common targets. Also, this work considers one of the multi-agent extensions of the POMDP framework which is the \textit{Decentralized Partially Observable Markov Decision Process (Dec-POMDP)}. This formulation is particularly suitable for decentralized execution in cluster satellite EO missions, where each agent determines its actions based solely on local observations \citep{oliehoek2016concise}. A Dec-POMDP is formally defined by the tuple:
\begin{equation}
    \mathcal{D}=\langle S, \{A_i\}_{i=1}^{N}, T, r, \{O_i\}_{i=1}^{N}, O, N, \gamma \rangle,
\end{equation}
where $S$ represents the set of environment states, and each agent $i$ has an action space $A_i$, forming the joint action space $A = A_1 \times \dots \times A_N$ for $N$ agents. The state transition function $T: S \times A \times S \rightarrow [0,1]$ describes the probability of transitioning from state $s$ to $s'$ given the joint action $\mathbf{a} = (a_1, \dots, a_N)$. The global reward function $r: S \times A \rightarrow \mathbb{R}$ provides feedback based on joint actions. Each agent $i$ has an observation space $O_i$, and the joint observation space is $O = O_1 \times \dots \times O_N$. The observation function $O: S \times A \times O \rightarrow [0,1]$ defines the probability of an agent receiving observation $o_i$ given state $s$ and joint action $\mathbf{a}$. The number of agents is $N$, and $\gamma \in [0,1]$ is the discount factor, which controls the importance of future rewards.

In a cooperative satellite cluster system, the global reward function (as in Eq. \ref{eq:reward}) is determined by unique image captures. This means that different satellites must capture different targets to maximize the total reward. Since Dec-POMDPs operate under decentralized execution, each satellite makes decisions based only on its partial observations while still contributing to the shared objective. This decentralized yet cooperative nature introduces challenges in coordinating image captures among satellites, as they must infer the actions of others without direct communication. Efficient policy learning in this setting requires strategies that promote diversity in target selection while ensuring optimal energy and data storage management.

\subsection{Heterogeneous Satellite Cluster Modelling}
In a heterogeneous satellite cluster, each satellite may differ in sensing payload, communication subsystem, energy capacity, storage availability, or orbital configuration. Such systems naturally lead to a heterogeneous multi-agent formulation, in which each satellite is modelled as an autonomous decision-maker with distinct capabilities and objectives. To capture these properties, the multi-satellite environment can be formalized as a stochastic game:
\begin{equation}
\mathcal{G} = \langle S, \{O_i\}, \{A_i\}, P, \{r_i\}, \gamma \rangle,
\end{equation}
where, $S$ denotes the global state space, $O_i$ and $A_i$ denote the local observation and action spaces of satellite $i$, $P(\cdot)$ is the state transition kernel, $r_i$ is the reward function associated with satellite $i$, and $\gamma$ is the discount factor. Due to heterogeneity, satellites do not share identical observation or action spaces, and thus $O_i \neq O_j$ and $A_i \neq A_j$ for some $i \neq j$. Each satellite selects actions using a decentralized policy.
\begin{equation}
\pi_{\theta_i}(a_i \mid o_i),
\end{equation}
which is parameterized by $\theta_i$ and conditioned only on its private observation. The joint policy is represented as a product of individual decentralized policies,
\begin{equation}
\Pi_{\boldsymbol{\theta}}(\mathbf{a} \mid \mathbf{o}) = \prod_{i=1}^N \pi_{\theta_i}(a_i \mid o_i),
\end{equation}
which reflects the decentralized execution setting commonly adopted in multi-agent reinforcement learning.

To explicitly characterize heterogeneity, each satellite $i$ can be associated with a capability tuple:
\begin{equation}
\mathcal{H}_i = \langle \mathcal{C}_i, \mathcal{E}_i, \mathcal{M}_i, \mathcal{S}_i \rangle,
\end{equation}
where $\mathcal{C}_i$ models the sensing or service capability of the satellite (e.g., optical or SAR payload), $\mathcal{E}_i$ represents energy and storage constraints, $\mathcal{M}_i$ defines mobility or kinematic limitations derived from orbital mechanics, and $\mathcal{S}_i$ captures sensing, communication, or coverage constraints. Heterogeneity implies that there exist agents $i$ and $j$ such that $\mathcal{H}_i \neq \mathcal{H}_j$, which in turn leads to differences in observation models, action feasibility sets, and policy behaviour. This relationship can be expressed formally as:
\begin{equation}
\mathcal{H}_i \neq \mathcal{H}_j \;\Longrightarrow\; (O_i \neq O_j)\ \lor\ (A_i \neq A_j)\ \lor\ (\pi_{\theta_i} \neq \pi_{\theta_j}).
\end{equation}

The presence of heterogeneous satellites also affects both the transition and reward structures of the environment. Because satellites differ in capability and subsystem limitations, the transition model becomes agent-dependent and can be expressed as:
\begin{equation}
P(s' \mid s, a_1, \dots, a_N) = f(\mathcal{H}_1, \dots, \mathcal{H}_N),
\end{equation}
where the heterogeneous capability profiles directly influence orbital evolution, task execution feasibility, energy consumption, and network connectivity. Likewise, the reward function is no longer symmetric, and in general $r_i(s, a_1, \dots, a_N) \neq r_j(s, a_1, \dots, a_N)$, since an imaging satellite, a relay satellite, and a task scheduler may each contribute differently to mission objectives.

This heterogeneity introduces significant challenges for multi-agent reinforcement learning in the stochastic game setting \citep{zhong2024heterogeneous}. First, the learning process becomes highly non-stationary, as agents with distinct policies update at different rates while influencing one another's state-transition dynamics. Second, asymmetric observation structures make coordination more difficult, since agents operate with fundamentally different and potentially incomplete information. Third, heterogeneous and unequal action spaces prevent the direct application of MARL methods originally designed for homogeneous agents with shared policies. Finally, credit assignment becomes more complex, as different satellites contribute unequally to global performance and must be evaluated under role-dependent reward structures. As a result, heterogeneous satellite clusters require specialized MARL algorithms capable of handling agent-specific policies, role-aware critics, and asymmetric information flows, making this an open and challenging research domain.

\subsection{Other Challenges}
\paragraph{1) Uncertainties and Randomness} 
A real-world system always contains uncertainties and randomness inherent in nature, arising from factors such as noise, disturbances, and unpredictable variations. In EO satellite missions, these uncertainties are particularly influenced by the availability of resources which affects the satellite's observation probability function and the necessity for taking actions. The initial condition before the learning started affects stability. For instance, under the shaded area, the satellite's policy must balance imaging action with the battery level. The other factor is the availability of initial data storage space, where the policy must balance imaging action with data downlinking to ensure there is enough memory space to capture more images. In other words, resources initial conditions directly impact observation quality and the feasibility of actions, making them critical in the POMDP model for optimal action planning in EO missions. Therefore, the POMDP tuple definition ($\mathcal{G}$) implicitly accommodates uncertainties and randomness, as the satellite cannot obtain complete state information at all times.

\paragraph{2) Non-Stationarity}
Non-stationarity in multi-satellite settings arises because the environment is constantly evolving as agents update their policies during training. Unlike single-agent scenarios, where the environment remains fixed, the presence of multiple agents creates a dynamic and unpredictable environment due to joint action dependencies. As one agent updates its policy, it alters the environment for others, causing the distribution of states and actions to shift. This violates the stationarity assumption in most reinforcement learning algorithms, leading to instability in training. Another challenge is \textit{reward interdependency}, where a satellite's reward is influenced by the actions of others. For example, the global reward based on unique image captures depends on the joint action vector $\mathbf{a} = (a_1, a_2, \dots, a_N)$. As satellites adjust their policies, the reward landscape changes, increasing the complexity of policy learning.

\subsection{Satellite Cluster Resources Optimization as Sequential Decision Making Process}
The objective of the satellites is to capture as many unique images as possible during their orbits. Previous works on single-satellite EO missions have formulated the problem as a sequential decision-making task in the well-known reinforcement learning framework, specifically as a Partially Observable Markov Decision Process (POMDP) \citep{stephenson2024reinforcement,stephenson2024using}. Building upon this, we formally define the multi-satellite EO mission as a Decentralized POMDP (Dec-POMDP) model, extending from the single-agent MDP framework.

\section{Multi-agent Solution Design}

\subsection{Reward Function Design}
The instantaneous reward $R_t$ integrates three mission objectives: data acquisition, resource utilization, and safe operation. The reward function is defined as:
\begin{equation}
R_t =
\begin{cases}
q_i - \rho_t + c_t, & \text{if an AoI is successfully captured}, \\
-\rho_t + \delta_t, & \text{if data are downlinked}, \\
-100, & \text{if a failure occurs}, \\
-\rho_t, & \text{if only power is consumed}, \\
0, & \text{otherwise},
\end{cases}
\label{eq:reward}
\end{equation}
where $q_i \in (0,1)$ represents the priority of the target AoI at time $t$.

\paragraph{1) Data Acquisition}
To maximize the collection of high-priority AoIs, the reward term $q_i$ promotes selecting 
targets with greater mission importance.

\paragraph{2) Resource Utilisation}
Resource efficiency is encouraged through three supporting terms:

$\bullet$ \textbf{Minimize power usage:}
    \begin{equation}
    \rho_t = \alpha \, \Delta Q_t \, (1 - Q_t), \qquad 
    \Delta Q_t = Q_{t-1} - Q_t,
    \end{equation}
    where $\rho_t$ penalizes excessive energy consumption ($\Delta Q_t$) times a constant $\alpha$.

$\bullet$ \textbf{maximize downlinking:}
    \begin{equation}
    \delta_t = \beta \, \Delta D_t, \qquad
    \Delta D_t = D_t - D_{t-1},
    \end{equation}
    which rewards successful transmission of collected data. The amount of the transferred data is calculated as $\Delta D_t$ multiply by a scalar constant ($\beta$).

$\bullet$ \textbf{maximize payload usage:}
    \begin{equation}
    c_t =
    \begin{cases}
    -1 + \sigma, & \sigma < 0.5 \ \text{and payload = SAR}, \\
    \sigma, & \sigma \ge 0.5 \ \text{and payload = SAR}, \\
    1 - \sigma, & \sigma < 0.5 \ \text{and payload = OPT}, \\
    -\sigma, & \sigma \ge 0.5 \ \text{and payload = OPT},
    \end{cases}
    \end{equation}
    where, $\sigma \in (0,1)$ is the cloud coverage ratio, ensuring SAR is used in cloudy 
    conditions and optical payloads are used in clear conditions.

\paragraph{3) Avoid Failure}
A large penalty is applied to discourage mission failure:
\begin{equation}
R_t = -100 \qquad \text{(failure penalty)}.
\end{equation}
If the satellite encounters a failure, a fault condition is triggered, represented as:
\begin{equation}
    Failure = (b_t < m_b \space \vee \space \text{any}(\hat{\Omega}\geq\Omega_{max})).
    \label{eq:failure}
\end{equation}
where, $m_b$ is the minimum battery level to trigger a failure. This ensures safe and continuous satellite operation.

\subsection{RL for Single Satellite}
To solve POMDP and Dec-POMDP of Autonomous EO Mission, a model-free approach is selected due to its flexibility to directly learn the policy, especially when the model of the system is complex and highly dynamic. Hence, RL is initially tailored to handle single-satellite EO mission then followed by the extension to MARL with multi-satellite constellation.

A single-satellite has its own on-board processor to execute the best autonomy policy. Here, the satellite is defined as an agent to solve the satellite decision making during performing EO Mission. RL is used to learn an optimal policy ($\pi^*$) which maps states to actions $\pi(a_t|s_t)$ and maximizes the expected cumulative reward, or return, over time \cite{sutton2018reinforcement}. The return at time step \( t \), denoted \( R_t \), is expressed as the sum of discounted rewards \( R_t = \sum_{i=0}^{\infty} \gamma^i r_{t+i+1}\). To measure how good an observation is under a particular policy, a State-Value function is defined as the expected cumulative reward obtained from Eq. \ref{eq:reward}: \(V_{\pi}(s)=\mathbb{E}_{\pi}\left[ R_t |S_t=s\right]\) and an Action-Value function: \(Q^\pi(s, a) = \mathbb{E}_\pi \left[ R_t | S_t = s, A_t = a \right]\).

\subsubsection{Proximal Policy Optimization}
Among the variety of the RL approaches, our work focus on the \textit{on-policy methods} which is Proximal Policy Optimization (PPO) \cite{schulman2017PPO}. This algorithm is a widely used reinforcement learning (RL) algorithm, particularly for solving high-dimensionality problems. It belongs to the family of policy optimization methods with actor-critic networks architecture, where the goal is to optimize the policy directly instead of learning a value function. PPO aims to improve the policy $\pi_{\theta}(a|s)$, parameterized by $\theta$, to maximize the expected cumulative reward: $J(\theta) = \mathbb{E}_{\tau \sim \pi_\theta} \left[ \sum_{t=0}^\infty \gamma^t r_t \right]$. Instead of directly optimizing $J(\theta)$, PPO uses a surrogate objective function to limit policy updates and stabilize the training phase:
\begin{equation}
    L^{\text{PPO}}(\theta) = \mathbb{E}_t \left[ \min\left(r_t(\theta) \hat{A}_t, \text{clip}(r_t(\theta), 1 - \epsilon, 1 + \epsilon) \hat{A}_t\right) \right],
\end{equation}
where $r_t(\theta)=\frac{\pi_{\theta}(a|s)}{\pi_{\theta_{old}}(a|s)}$ is the probability ratio between the new and old policies. $\hat{A}_t$ is the estimated advantage function, typically calculated using the Generalized Advantage Estimation (GAE). And $\epsilon$ is a hyper-parameter controlling the clipping range. The "clip" term ensures that the probability ratio $r_t(\theta)$ does not deviate too much from 1, avoiding overly large policy updates. This stabilizes training and prevents performance collapse, which is more suitable for a realistic satellite simulation with uncertainty and randomness. 

PPO also includes a value function loss to improve the policy's state value predictions: $L^{\text{value}}(\theta) = \mathbb{E}_t \left[ \left( V_\theta(s_t) - V_t^\text{target} \right)^2 \right]$. Then, to encourage exploration, an entropy bonus term is added: $L^{\text{entropy}}(\theta) = \mathbb{E}_t \left[ -\pi_\theta(a_t | s_t) \log \pi_\theta(a_t | s_t) \right]$. The total loss function for PPO is a combination of the surrogate objective, value function loss, and entropy bonus: $L^{\text{total}}(\theta) = L^{\text{PPO}}(\theta) - c_1 L^{\text{value}}(\theta) + c_2 L^{\text{entropy}}(\theta)$, where $c_1$ and $c_2$ are the coefficients used for balancing the contributions of the different terms.

\subsection{MARL with CTDE Framework Algorithms}
The centralized training with decentralized execution (CTDE) framework has been proven to be an effective paradigm in multi-agent reinforcement learning (MARL), as it enables agents to learn coordinated behaviours while maintaining independent decision-making during deployment. The following subsection discusses how the CTDE principle can be adapted to model and train heterogeneous satellite clusters, where each satellite learns a distinct policy under a shared mission objective while benefiting from centralized information during training.

\subsubsection{CTDE for Heterogeneous Satellite Cluster}
\begin{figure}[t]
    \centering
    \includegraphics[width=0.7\linewidth]{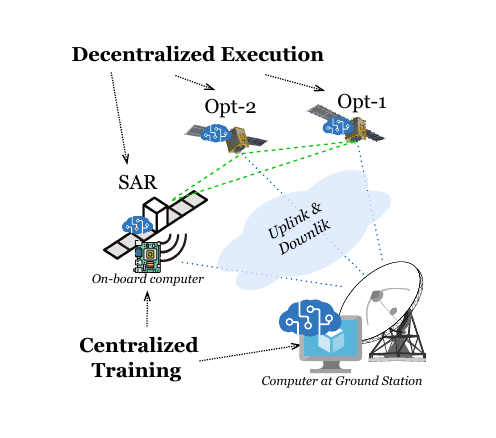}
    \caption{Centralized Training, Decentralized Execution (CTDE) learning frameworks for satellite cluster EO missions. This framework balances both approaches, keeping training centralized while allowing satellites to execute independently, reducing the need for real-time communication except during training and data downlinking.}
    \label{fig:CTDE}
\end{figure}

 In Centralized Training Decentralized Execution (CTDE) framework, agents are trained with centralized knowledge, such as joint states, actions, and rewards to capture all agent's behaviour, but they execute their policies independently based on local observations (see Fig. \ref{fig:CTDE}. This feature enables better scalability and is applicable to the satellite's on-board computer in decentralized mode. Each agent $i \in \mathcal{N}$ learns a policy $\pi_i(a_i|o_i)$, where $o_i$ is the local observation of agent-$i$. The training uses a centralized critic $Q_{\theta}(o,\boldsymbol{a})$ that evaluates joint actions based on the global observation $\boldsymbol{o}$: $Q_\theta(s, \mathbf{a}) = \mathbb{E} \left[ \sum_{t=0}^\infty \gamma^t r(s_t, \mathbf{a}_t) \mid s_0 = s, \mathbf{a}_0 = \mathbf{a} \right]$. And during execution, the policy of each agent is still in a decentralized manner: $\pi_i(a_i|o_i)$. The CTDE enjoys the advantage intersection of both fully centralized and decentralized framework. It strikes a balance between leveraging centralized information during training and enabling decentralized decision-making during execution.  

In our works, three recent state-of-the-art methods under CTDE framework have been selected to be compared: 

\subsubsection{Multi-agent PPO (MAPPO)}
MAPPO is an extension of PPO designed specifically for multi-agent systems \cite{yu2022surprising}. It incorporates centralized critics and decentralized policies to improve performance in MARL tasks. MAPPO uses a single centralized critic shared by all agents, allowing the evaluation of the global state to stabilize learning and mitigate non-stationarity: $V_i^{\text{centralized}}(s) \approx \mathbb{E} \left[ \sum_{t=0}^\infty \gamma^t r_{i,t} \mid s_0 = s \right]$, where \( s \) represents the global state, \( \gamma \) is the discount factor, and \( r_t \) is the reward at time step \( t \). The loss function for the policy optimization in MAPPO is given by: 
\begin{equation}
    L^{\text{MAPPO}}(\theta_i) = \mathbb{E}_t \left[ \min\left(r_t(\theta_i) \hat{A}_t, \text{clip}(r_t(\theta_i), 1-\epsilon, 1+\epsilon) \hat{A}_t\right) \right],
\end{equation}
where: \( r_t(\theta_i) = \frac{\pi_{\theta_i}(a_t \mid o_t)}{\pi_{\theta_i}^{\text{old}}(a_t \mid o_t)} \) is the probability ratio,\( \hat{A}_t \) is the global advantage function, and \( \epsilon \) is the clipping parameter.

\subsubsection{Heterogeneous Agent Trust Region Policy Optimization (HATRPO)}
Consider a cooperative stochastic game / Dec-POMDP with $N$ agents and discount factor $\gamma$. Using the importance-sampled surrogate objective in the TRPO style, the per-agent surrogate objective (estimated from trajectories collected with the behaviour policy $\pi_{\theta^{\text{old}}}$) is:
\begin{equation}
  \mathcal{L}_{\theta_i}^{\text{sur}}(\theta_i)
  = \mathbb{E}_{s,o_i,a_i \sim \mathcal{D}}
  \left[
    \frac{\pi_{\theta_i}(a_i\mid o_i)}{\pi_{\theta_i^{\text{old}}}(a_i\mid o_i)} \; \hat{A}_i(s,\mathbf{a})
  \right],
\end{equation}
where $\hat{A}_i(s,\mathbf{a})$ is the advantage estimate for agent $i$ (see GAE below) and $\mathcal{D}$ denotes empirical state–action samples.

HATRPO uses the trust-region constraint (per agent) to ensure stable update and enforces a KL constraint per agent:
\begin{equation}
  \mathbb{E}_{o_i\sim\mathcal{D}}\Big[
    \mathrm{KL}\big(\pi_{\theta_i^{\text{old}}}(\cdot\mid o_i)\;\|\;\pi_{\theta_i}(\cdot\mid o_i)\big)
  \Big] \le \delta,
\end{equation}
with trust-region radius $\delta>0$.

This algorithm has a constrained update for trust region optimization for agent $i$ and is designed as:
\begin{align}
  \theta_i^{\text{new}} &= \arg\max_{\theta_i} \ \mathcal{L}_{\theta_i}^{\text{sur}}(\theta_i)
  \quad\text{s.t.}\quad
  \mathbb{E}_{o_i}\big[ \mathrm{KL}(\pi_{\theta_i^{\text{old}}}\| \pi_{\theta_i}) \big] \le \delta.
\end{align}
Linearizing the objective and quadratically approximating the KL leads to the natural-gradient step:
\begin{align}
  g_i &:= \nabla_{\theta_i} \mathcal{L}_{\theta_i}^{\text{sur}}(\theta_i)\Big|_{\theta_i=\theta_i^{\text{old}}}, \\
  H_i &:= \nabla_{\theta_i}^2 \; \mathbb{E}_{o_i}\big[ \mathrm{KL}(\pi_{\theta_i^{\text{old}}}\| \pi_{\theta_i})\big]\Big|_{\theta_i=\theta_i^{\text{old}}}
\end{align}
(where $H_i$ is the Fisher information matrix under the old policy). The unconstrained natural-gradient direction is $p_i = H_i^{-1} g_i$. Scaling $p_i$ to satisfy the quadratic KL constraint yields the final step:
\begin{equation}
  \theta_i^{\text{new}} = \theta_i^{\text{old}} + \alpha_i \, p_i,
  \quad\text{with}\quad
  \alpha_i = \sqrt{\frac{2\delta}{\,g_i^\top H_i^{-1} g_i\,}}.
\end{equation}
In practice $H_i^{-1} g_i$ is computed approximately via conjugate-gradient and $g_i^\top H_i^{-1} g_i$ by a Fisher-vector product.

Similar to PPO, HATRPO commonly uses GAE computed with the centralized critic $V_\phi(s)$, and the per-agent advantage $\hat{A}_i(s_t,\mathbf{a}_t)$ can be derived from the shared value and reward decomposition used in the environment (e.g., same $\hat{A}_t$ for fully cooperative rewards or agent-specific terms for heterogeneous rewards).

The centralized value function is trained to minimize the squared error to empirical returns or GAE-based targets. A typical critic loss is:
\begin{equation}
  \mathcal{L}_V(\phi) \;=\; \mathbb{E}_{s_t\sim\mathcal{D}}
  \Big[ \big( V_\phi(s_t) - \hat{V}_t \big)^2 \Big],
\end{equation}
where $\hat{V}_t$ is a target return (e.g., $\hat{V}_t = \hat{A}_t + V_\phi(s_t)$ or the GAE-return estimate). The critic is updated by gradient descent on $\mathcal{L}_V(\phi)$.

HATRPO retains TRPO's stability benefits while allowing each agent to have distinct observation/action spaces and separate policy parameters. In implementation, per-agent KL constraints, conjugate-gradient solves, and efficient Fisher-vector products are essential for scalability.

\subsubsection{Heterogeneous Agent PPO (HAPPO)} 
This algorithm extends MAPPO by accounting for heterogeneous agents with distinct state-action spaces or roles and the sequential update scheme \cite{zhong2024heterogeneous}. It uses individual advantage functions and decentralized policies while maintaining centralized critics.
In HAPPO, the centralized value function is agent-specific to handle heterogeneous agents: $V_i^{\text{centralized}}(s) \approx \mathbb{E} \left[ \sum_{t=0}^\infty \gamma^t r_{i,t} \mid s_0 = s \right]$, where \( i \) denotes the agent index and \( r_{i,t} \) is the reward specific to agent \( i \). And, the loss function for HAPPO is denoted by:
\begin{equation}
    L_i^{\text{HAPPO}}(\theta_i) = \mathbb{E}_t \left[ \min\left(r_{i,t}(\theta_i) \hat{A}_{i,t}, \text{clip}(r_{i,t}(\theta_i), 1-\epsilon, 1+\epsilon) \hat{A}_{i,t}\right) \right],
\end{equation}
where \( r_{i,t}(\theta_i) = \frac{\pi_{\theta_i}(a_{i,t} \mid o_{i,t})}{\pi_{\theta_i}^{\text{old}}(a_{i,t} \mid o_{i,t})} \) and \( \hat{A}_{i,t} \) are the advantage functions of agent \( i \).

All algorithms, MAPPO, HATRPO, and HAPPO operate effectively in the CTDE paradigm. It has main differentiations in the centralized critics, where MAPPO uses a single global critic for shared evaluation, while HATRPO and HAPPO incorporate agent-specific critics for greater flexibility in heterogeneous systems. It shares the same decentralized execution settings, where each agent executes its policy based solely on local observations, making these algorithms suitable for real-world scenarios where the global state information is unavailable during execution.

\section{Experimental Results}

\subsection{Environment Settings}
\subsubsection{Heterogeneous Cluster Orbital Settings}
The orbital parameters in BSK-RL used for our experiments are shown in Table \ref{tab:orbitalParams}.

\begin{table}[h]
\centering
\caption{Orbital Parameters}
\label{tab:orbitalParams}
\begin{tabular}{lcc}
\hline
\multicolumn{1}{c}{\textbf{Parameter}} & \textbf{Single-Sat.} & \textbf{Cluster [OPT-1, OPT-2, SAR]} \\ \hline
Inclination   (deg.)                 & 40                     & [41, 41, 40]               \\
Offset (deg.)                        & -75                    & [-74, -74, -75]            \\
Num. of   Planes                     & 1                      & 3                \\
Altitude (km)                        & 500                    & 500              \\ \hline
\end{tabular}
\end{table}

\subsubsection{Satellite Parameters}
Several satellite parameters that define the satellite specifications are adjustable from the BSK-RL's environment definition, as shown in Table \ref{tab:satParams}.

\begin{table}[H]
\centering
\caption{Satellite Default Parameters}
\label{tab:satParams}
\begin{tabular}{lcc}
\hline
\multicolumn{1}{c}{\textbf{Parameter Name}} & \textbf{Default Value} & \textbf{Unit} \\ \hline
\textbf{Data Storage   and Transmission:}   & \textbf{}              & \textbf{}     \\ \cline{1-1}
Data Storage   Capactity                    & 500                    & GB            \\
Initial Data   Storage Level                & 0                      & \%            \\
Instrument   Baud Rate                      & 500                   & kbps          \\
Transmitter   Baud Rate                     & 100                    & Mbps          \\ \hline
\textbf{Power System:}                      &                        &               \\ \cline{1-1}
Battery Capacity                  & 400                    & W.h           \\
Initial   Battery Level                     & 100                    & \%            \\
Solar Panel   Area                          & 1                      & $m^2$         \\
Solar Panel   Efficiency                    & 20                     & \%            \\
Base Power   Draw                           & -10                    & W             \\
Instrument   Power Draw                     & -30                    & W             \\
Thruster Power   Draw                       & -80                    & W             \\ \hline
\textbf{Satellite   Attitude:}              &                        &               \\ \cline{1-1}
Image Attitude   Error                      & 0.1                    & Degree        \\
Image Rate   Error                          & 0.1                    & Degree        \\
Attitude   Control Input Max.               & 0.4                    & -             \\
Attitude Rate   Control Input Max.          & 0.1                    & -             \\
Servo P   constant                          & 30                    & -             \\
Servo Ki   constant                         & 5.0                      & -             \\
Disturbance   vector                        & {[}0, 0, 0{]}          & -             \\
Max. Reaction   Wheel speed                 & 6000                   & RPM           \\
Initial   Reaction Wheel speed              & {[}0, 0, 0{]}          & RPM           \\ \hline
\end{tabular}
\end{table}

\subsubsection{Area of Interests (AoIs)}
For all scenarios, the AoIs are defined alongside the satellite orbit ground track. It consists of 11 different regions and for each regions are divided into smaller AoIs with $3\times15$ AoIs arranged in grids. Different regions are assumed have different cloud cover conditions and priorities and these values are pre-defined and can be customized from a .csv file based on the user requirements or customer needs.

\begin{figure}[t]
    \centering
    \includegraphics[width=0.5\linewidth]{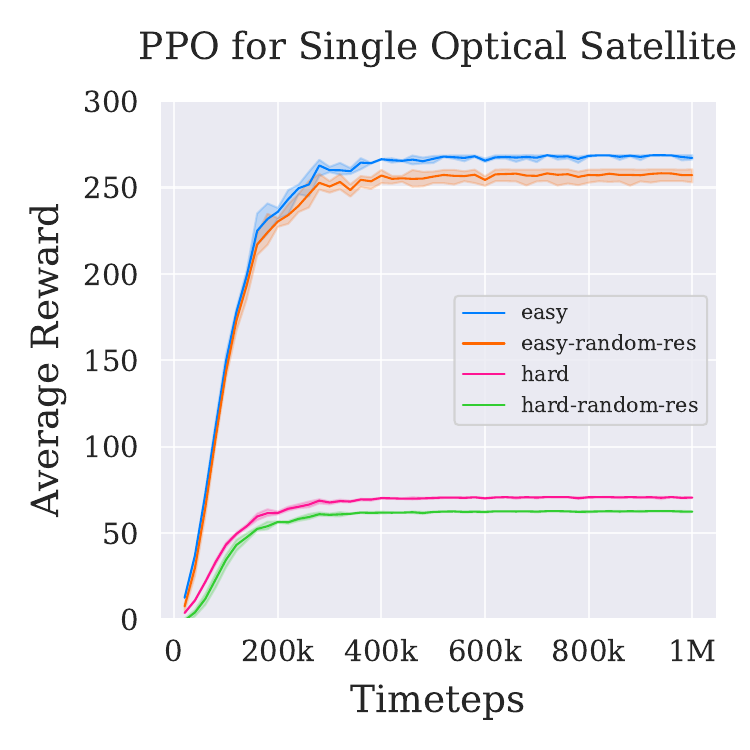}
    \caption{Single Optical Satellite learning curve in different cases.}
    \label{fig:single-sat}
\end{figure}

\subsubsection{Ground Stations}
We refer to the commercially available ground station service provided by Amazon Web Services (AWS) Inc. \citep{aws2025groundstation}. Based on this service, there are 12 ground stations used in our simulation. The coordinates are defined in the simulator by following the antenna locations distributed in different cities around the world. 

\subsubsection{Cluster Resources Scenarios}
\paragraph{1) \textbf{easy}}
This scenario is designed based on default satellite parameters and all of resources initial conditions are set to be fully available for executing a mission. The battery level is fully charged (100\%), memory storage is empty (0\%), there is no randomness at the attitude disturbance and reaction wheel speed initialization). It simulates the mission with less challenge of the cluster resources availability.

\paragraph{2) \textbf{easy-random-res}}
In this scenario, the parameters are identical to the easy scenario, while the initialization of battery, memory storage, disturbance and reaction wheel speed are randomized. This adds more challenge to the cluster to perform the EO mission and evaluate the generalization performance of the policy across different resources initial condition. This scenario simulates more realistic case of the real satellite mission, where the resources initial condition may vary depends on the states of its previous mission. The battery level is initialized randomly between 80-95\%, the memory storage is 60-80\%, disturbance is randomized with normal distribution with the scale=$10^{-4}$, and the reaction wheels are uniformly randomized between -3000 to 3000 RPM.

\paragraph{3) \textbf{hard}}
The \textit{hard} scenario is designed to simulate the cluster under conditions of restricted resources. The downlink transmission speed (baud-rate) is assumed to degrade up to 30\% from its default value. Additionally, the battery level is directly initialized as 85\% and the memory storage is initialized as 60\%. This scenario introduce more difficulties especially to balance the memory storage utilization, if too many data collected the memory will be full and can not be properly downlinked due to the issue of the transmission speed.

\paragraph{4) \textbf{hard-random-res}}
This scenario integrates the \textit{hard} scenario with randomness factor that are identical to the \textit{easy-random-res}. In other words, all of the randomness parameters are set equal to the \textit{easy-random-res} scenario settings, except the downlink transmission, yet speed, battery and memory initialization are randomized. Therefore, this scenario demonstrates more realistic challenge of the resource availability as well as its randomness nature at the same time.

\subsection{Policy Training Results}
In our experiment, the policies of different MARL algorithms are trained under the same computer with Intel(R) Core(TM) i9-14900K CPU, 32 GB RAM, and 24 GB GPU NVIDIA GeForce RTX 4090. The training is terminated at 1 million training time steps with 20 parallel-processing environments (rollouts) and it took around 3-4 hours for each seed (3 different seeds).

\subsubsection{Single Satellite}
Firstly, a single agent PPO algorithm is evaluated for one satellite resource optimization problem. An optical satellite is deployed in the simulator to perform an EO mission. Different cases are used to obtain preliminary mission simulation results of the pre-defined mission at different difficulty levels. Overall, the results of the easy scenario give higher rewards compared to the hard scenario. This ensures our design of the reward function that consists of resource optimization and mission completion. The presence of randomness is also an important part to be seen from this single satellite setting. As shown in the Fig. \ref{fig:single-sat}, the single satellite configuration represents the basic scenario in which one spacecraft must manage sensing, power, and downlink decisions independently. The results show that PPO performs reliably in the easy and easy-random-resource scenarios, achieving relatively high average returns, while its performance drops significantly in the hard and hard-random-resource settings. This indicates that a single satellite struggles to handle operational uncertainty and resource variability, particularly when the environment becomes complex or resource-constrained.

The evaluation after 1M training time steps are provided in Table \ref{tab:single_sat_results}. 

\begin{table}[t]
\centering
\caption{Single Satellite PPO Performance Across Scenarios}
\begin{tabular}{l c}
\toprule
\textbf{Scenario} & \textbf{PPO (Avg $\pm$ Std)} \\ \midrule
easy              & $268.68 \pm 0.00$ \\ 
easy-random-res   & $258.25 \pm 3.19$ \\ 
hard              & $70.93 \pm 0.00$ \\ 
hard-random-res   & $62.80 \pm 0.07$ \\ 
\bottomrule
\end{tabular}
\label{tab:single_sat_results}
\end{table}

\begin{figure}[t]
    \centering
    \includegraphics[width=0.8\linewidth]{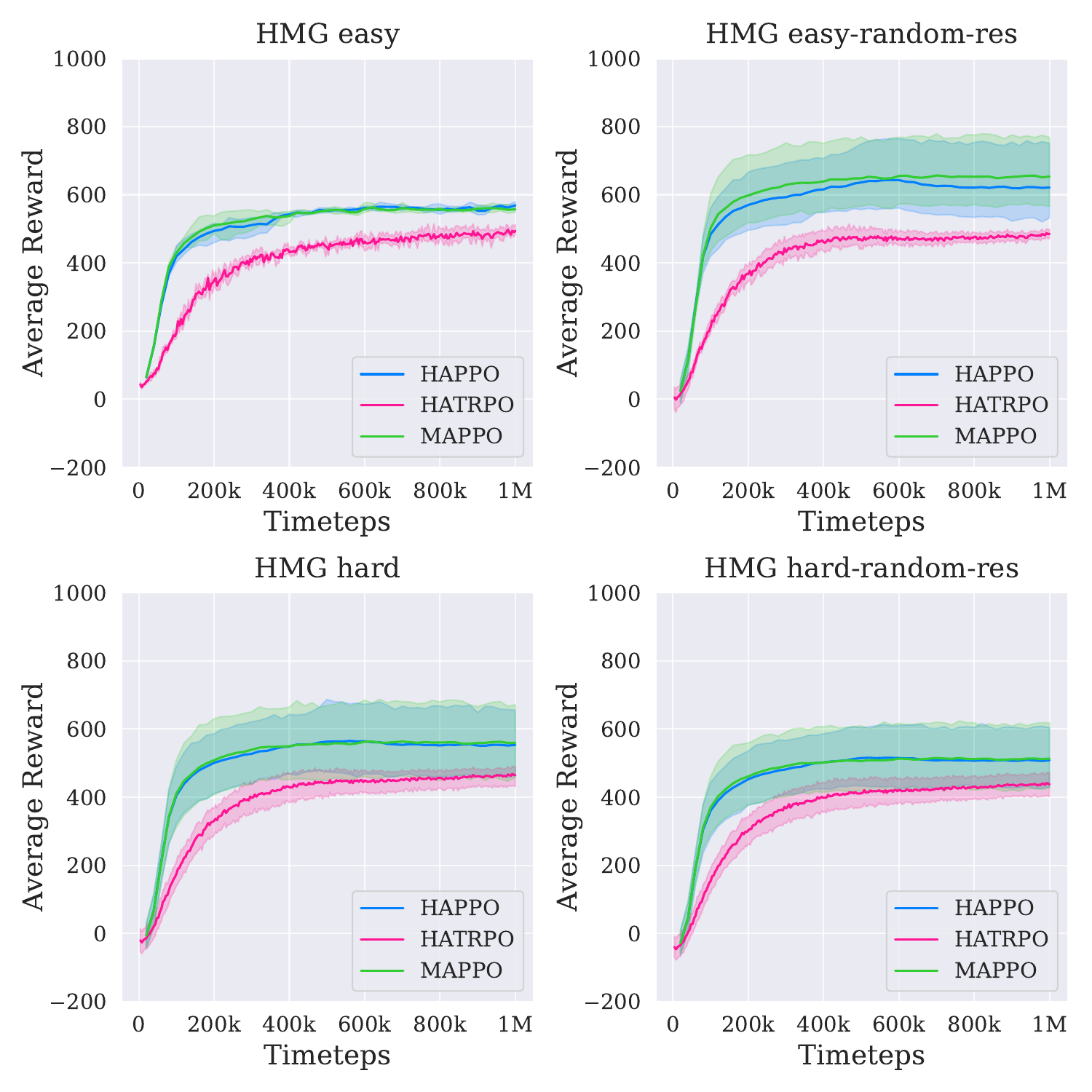}
    \caption{Homogeneous  Satellite Cluster (3-Optical)}
    \label{fig:hmg_training}
\end{figure}

\begin{figure}[t]
    \centering
    \includegraphics[width=0.8\linewidth]{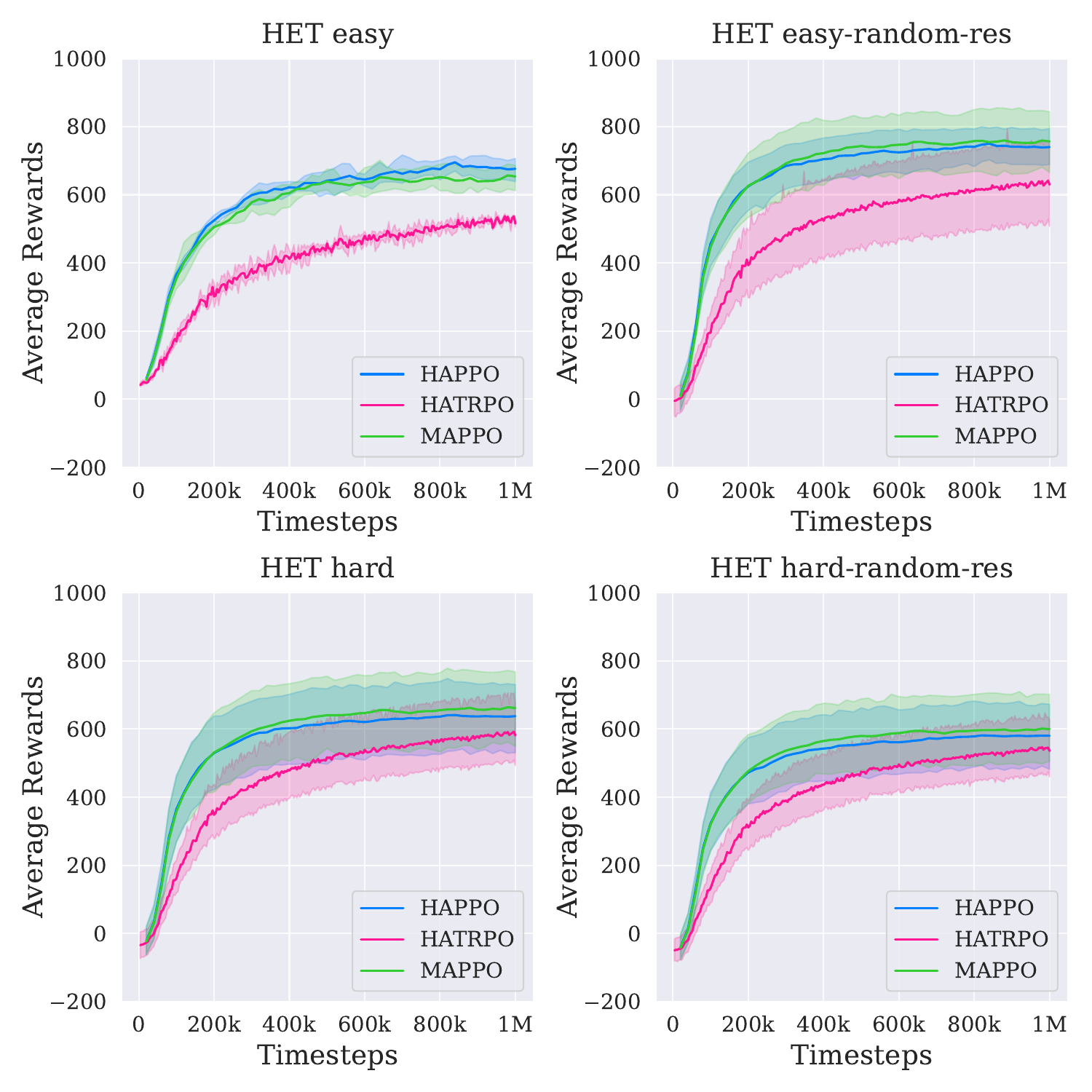}
    \caption{Heterogeneous Satellite Cluster (2-Optical and 1-SAR)}
    \label{fig:het_training}
\end{figure}

\subsubsection{Performance Evaluation in Homogeneous Satellite Cluster}
The homogeneous satellite cluster demonstrates a clear improvement in performance over the single satellite setup across all scenarios (see Fig \ref{fig:hmg_training}. With multiple identical agents sharing comparable sensing and downlink capabilities, cooperative behaviour emerges naturally, leading to more efficient task allocation and higher cumulative rewards. As shown in Table \ref{tab:homogeneousPerformance}, MAPPO and HAPPO shows stability learning performance in easy scenario, even with the presence of randomness, indicating that centralized training with decentralized execution is effective in coordinated cluster operations. The homogeneity of the agents simplifies policy learning and results in more consistent behaviour, making it a robust and scalable configuration for constellation-level resource allocation tasks. In hard scenarios, HATRPO performs better than the others, this indicates the importance of heterogeneous role between agents under restricted resource condition.

\begin{table}[t]
\centering
\caption{Performance of MAPPO in the Homogeneous Cluster configuration.}
\begin{tabular}{lccc}
\toprule
\textbf{Scenario} & \textbf{MAPPO} & \textbf{HATRPO} & \textbf{HAPPO} \\
\midrule
easy              & 559.45 $\pm$ 19.51   & 548.37 $\pm$ 12.49 & \textbf{561.97 $\pm$ 14.55} \\
easy-random-res   & \textbf{744.32 $\pm$ 137.34}  & 495.96 $\pm$ 16.72 & 656.32 $\pm$ 198.0696 \\
hard              & 372.59 $\pm$ 4.21  & \textbf{461.80 $\pm$ 50.64} & 424.02 $\pm$ 62.55 \\
hard-random-res   & 381.98 $\pm$ 38.63 & \textbf{390.95 $\pm$ 22.70} & 382.00 $\pm$ 16.95  \\
\bottomrule
\end{tabular}
\label{tab:homogeneousPerformance}
\end{table}

Table \ref{tab:homogeneousPerformance} presents the comparative evaluation results of MAPPO, HATRPO, and HAPPO in the homogeneous satellite cluster configuration, where all agents possess identical sensing and communication capabilities. This experiment assesses whether heterogeneous-agent MARL frameworks retain their effectiveness and coordination stability even when agent specialization is absent (identical).

\paragraph{1) Easy Scenario}
In the easy scenario, all algorithms achieve similar performance, with HAPPO slightly outperforming the others (561.97 ± 14.55), followed by MAPPO (559.45 ± 19.51) and HATRPO (548.37 ± 12.49). The narrow reward differences and low variance indicate that all three heterogeneous-agent algorithms can maintain consistent and balanced cooperation even in a symmetric agent setup. This confirms that the heterogeneous policy structures do not hinder learning in homogeneous environments.

\paragraph{2) Easy-Random-Res Scenario}
When resource randomness is introduced, MAPPO attains the highest reward (744.32 ± 137.34), demonstrating robust adaptability to stochastic initial resource conditions. HAPPO also performs competitively (656.32 ± 198.06) but with higher variance, suggesting occasional instability in convergence. HATRPO, while more stable, achieves lower average performance (495.96 ± 16.72). These results suggest that the centralized critic in MAPPO provides effective gradient stabilization across agents even when resource conditions fluctuate, reinforcing its scalability under homogeneous resource uncertainty.

\paragraph{3) Hard Scenario}
Under the constrained hard condition, HATRPO leads with 461.80 ± 50.64, outperforming both HAPPO (424.02 ± 62.55) and MAPPO (372.59 ± 4.21). This confirms the advantage of the trust-region optimization mechanism in maintaining stability under limited energy and memory. The superior performance of HATRPO again aligns with its robust policy regulation, ensuring agents learn consistent cooperative strategies without aggressive updates that could destabilize training.

\paragraph{4) Hard-Random-Res Scenario}
In the most challenging hard-random-res case, all three algorithms achieve comparable performance, with HATRPO marginally superior (390.95 ± 22.70) to MAPPO (381.98 ± 38.63) and HAPPO (382.00 ± 16.95). Despite the severe stochasticity, all algorithms maintain reasonable average rewards, confirming the generalization ability of heterogeneous-agent MARL designs even in uniform agent settings.

The results across all homogeneous scenarios reveal that heterogeneous-agent MARL frameworks remain highly effective and stable even when applied to homogeneous satellite clusters. Their shared centralized learning and decentralized execution paradigms enable agents to coordinate efficiently, regardless of whether they possess distinct or identical capabilities. Although MAPPO has worst average reward, it demonstrates strong adaptability to the other scenarios, benefiting from a shared critic that promotes coordinated learning among identical agents. HATRPO maintains superior robustness in constrained environments due to its controlled policy update mechanism. HAPPO shows consistently competitive results in easier scenarios, validating its ability to generalize across both heterogeneous and homogeneous domains.

\textbf{Insight of this finding:} The Heterogeneous-agent MARL frameworks are not limited to heterogeneous systems. Instead, they can naturally adapt to homogeneous system without losing efficiency or stability. This property makes them universally applicable for both mixed and uniform satellite clusters, providing a unified solution framework for multi-satellite coordination problems under diverse mission and resource conditions.

\subsubsection{Performance Evaluation in Heterogeneous Satellite Cluster}
The heterogeneous satellite cluster achieves the best overall mission performance, highlighting the benefits of complementary sensing (SAR and Optical sensors) and resource capabilities across agents. The results show that MAPPO, HATRPO, and HAPPO each leverage heterogeneity to improve task coverage, adaptability, and robustness, particularly in dynamic or resource-limited scenarios (see Fig. \ref{fig:het_training}). By enabling specialized agents (e.g., optical vs. SAR payloads) to allocate tasks according to their strengths, the constellation improves both data acquisition and downlink efficiency. Although learning becomes more complex due to differing observation and action spaces, the heterogeneous cluster ultimately delivers superior learning stability and better robustness to the presence of randomness.

\begin{table}[t]
\label{tab:heterogeneousPerformance}
\centering
\caption{Performance comparison of MAPPO, HATRPO, and HAPPO in the Heterogeneous Cluster configuration.}
\setlength{\tabcolsep}{4pt}
\renewcommand{\arraystretch}{1.1}
\begin{tabular}{lccc}
\toprule
\textbf{Scenario} & \textbf{MAPPO} & \textbf{HATRPO} & \textbf{HAPPO} \\
\midrule
easy            & 623.55 $\pm$ 17.15 & 585.27 $\pm$ 21.71 & \textbf{684.07 $\pm$ 40.47} \\
easy-random-res & \textbf{875.73 $\pm$ 55.28} & 790.15 $\pm$ 169.72 & 806.13 $\pm$ 22.81 \\
hard            & 496.38 $\pm$ 70.28 & \textbf{598.42 $\pm$ 16.82} & 454.49 $\pm$ 24.22 \\
hard-random-res & 410.05 $\pm$ 46.70 & \textbf{483.11 $\pm$ 84.89} & 411.90 $\pm$ 42.45 \\
\bottomrule
\end{tabular}
\label{tab:heterogeneous_cluster}
\end{table}

The Table \ref{tab:heterogeneous_cluster} summarizes the quantitative performance of three MARL algorithms (MAPPO, HATRPO, and HAPPO) under four different heterogeneous cluster scenarios: easy, easy-random-res, hard, and hard-random-res. The results reflect the algorithms’ capability to manage complex multi-satellite cooperation tasks involving heterogeneous resource constraints, stochasticity, and dynamic mission conditions.

\paragraph{1) Easy Scenario}
In the easy scenario, HAPPO achieves the highest mean reward (684.07 $\pm$ 40.47), outperforming both MAPPO (623.55 $\pm$ 17.15) and HATRPO (585.27 $\pm$ 21.71). This indicates that the heterogeneous actor–critic design of HAPPO allows better adaptation to distinct agent roles when the environment is relatively stable and resources are plentiful. The low variance in all methods shows that policy convergence is stable and the environment poses limited stochastic challenges.

\paragraph{2) Easy-Random-Res Scenario}
When resource randomness is introduced, performance differences become more pronounced. MAPPO obtains the best score (875.73 $\pm$ 55.28), suggesting superior robustness in handling moderate stochasticity in resource states, likely due to its centralized critic stabilizing policy updates across agents. HAPPO (806.13 $\pm$ 22.81) and HATRPO (790.15 $\pm$ 169.72) perform slightly lower, with HATRPO showing higher variance, implying sensitivity to dynamic or randomly initialized resource conditions.

\paragraph{3) Hard Scenario}
Under the more constrained and challenging hard scenario, HATRPO demonstrates clear superiority (598.42 $\pm$ 16.82), outperforming MAPPO (496.38 $\pm$ 70.28) and HAPPO (454.49 $\pm$ 24.22). This highlights HATRPO’s strength in trust-region–based updates, which provide controlled policy adaptation and stability when resources are limited or competition among agents increases. The consistent performance also aligns with the previously observed robust cooperative behaviour under resource stress (as shown in the policy execution snapshot).

\paragraph{4) Hard-Random-Res Scenario}
The hard-random-res case, combining severe resource constraints with stochastic initial states, presents the most difficult environment. Again, HATRPO achieves the best performance (483.11 $\pm$ 84.89), outperforming MAPPO (410.05 $\pm$ 46.70) and HAPPO (411.90 $\pm$ 42.45). Although the variance increases due to environmental randomness, HATRPO maintains the highest resilience, confirming its effectiveness for heterogeneous and uncertain multi-satellite missions.

Overall, the results illustrate a clear complementarity among the three algorithms. HAPPO excels in structured and stable environments, where fine-grained actor adaptation dominates. MAPPO achieves strong performance when moderate randomness is introduced, benefiting from its proximal update stability. HATRPO performs best under severe and random conditions, balancing learning stability and adaptability through its trust-region optimization mechanism. 

\textbf{Insight of this finding:} These outcomes validate the importance of algorithmic robustness in heterogeneous multi-agent systems and demonstrate how HATRPO effectively generalizes across challenging scenarios by maintaining policy stability under constrained resource dynamics.

\subsection{Training Robustness Performance to Initial Resources Condition}
\subsubsection{Battery Level Initial Condition Robustness Experiment}
\begin{figure}[t]
    \centering
    \includegraphics[width=\linewidth]{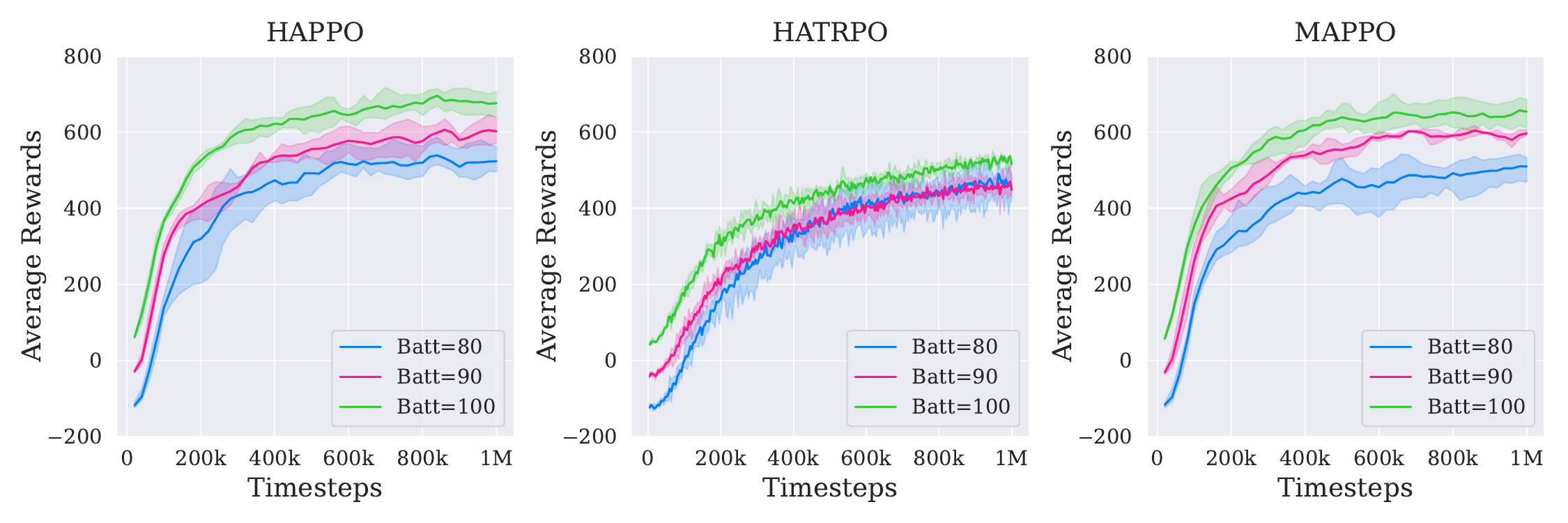}
    \caption{Training Robustness performance acrros different Battery resource initial condition}
    \label{fig:robustness-batt}
\end{figure}

This experiment evaluates the effect of different initial battery capacities (80\%, 90\%, and 100\%) on the training performance of three heterogeneous MARL algorithms (see Fig. \ref{fig:robustness-batt}), while the other parameters are set in their default value. HAPPO shows a consistent trend across battery conditions, with all curves converging to similar average reward levels. The slight differences during early time steps indicate minor sensitivity to initial energy levels, but convergence demonstrates strong adaptability and robustness to varying battery states. HATRPO exhibits more noticeable fluctuations in the early phase of training. While the final performance remains comparable among battery levels, the convergence rate is somewhat slower for the lower initial battery (80\%), suggesting a moderate dependence on the initial energy state during early exploration. MAPPO maintains the most stable and uniform performance across all battery settings. The convergence patterns overlap substantially, confirming MAPPO’s strong robustness and energy-condition invariance during learning.

\begin{figure}[h]
    \centering
    \includegraphics[width=\linewidth]{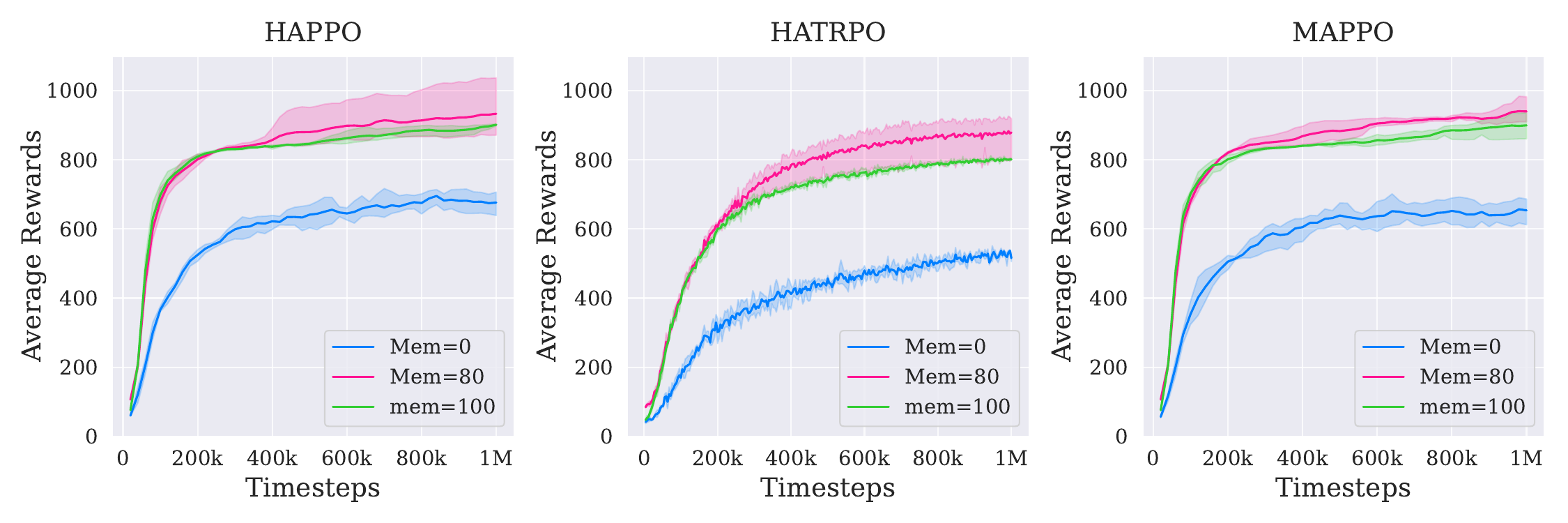}
    \caption{Training Robustness performance acrros different Memory resource initial condition}
    \label{fig:robustness-mem}
\end{figure}

\subsubsection{Memory Storage Initial Condition Robustness Experiment}
This experiment investigates the robustness of MARL algorithms under different initial memory storage conditions (0\%, 80\%, and 100\%) as shown in Fig. \ref{fig:robustness-mem} and keeps the other parameters as default value. HAPPO achieves high and consistent performance across all memory states, with minimal divergence between curves. At 0\% initial memory, learning proceeds efficiently, indicating HAPPO’s ability to adapt to resource scarcity without degradation in final performance. HATRPO shows slightly larger variance between conditions in early training, especially at 0\% memory. However, all settings converge to similar average reward levels, reflecting reliable long-term robustness despite early instability. MAPPO demonstrates the most stable and rapid convergence behaviour. The reward curves are nearly overlapping across all memory levels, signifying strong invariance and efficient policy adaptation regardless of initial resource conditions. Overall, the training robustness with respect to memory initialization is well maintained for all algorithms. MAPPO shows the highest consistency and stability, HAPPO follows closely, and HATRPO exhibits minor sensitivity during early exploration.

\begin{figure}[t]
    \centering
    \includegraphics[width=\linewidth]{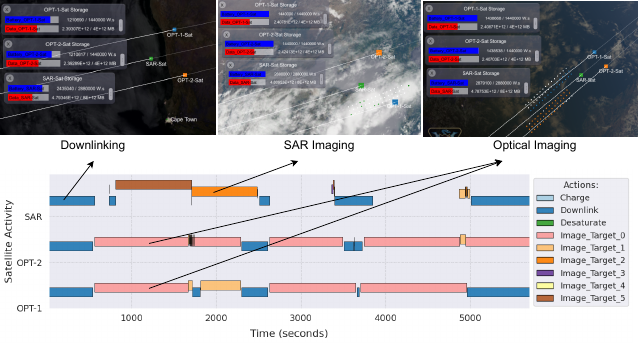}
    \caption{Satellite activity snapshot of HATRPO policy in \textit{hard} scenario. After training phase, the policies are deployed to the cluster and each satellite demonstrates different strategy to complete the mission in one orbit. The downlink action is selected once the satellite has opportunity to a ground station and no AoI around it. The SAR satellite captures the AoI with high cloud coverage. And both Optical satellites capture different region of AoI close to its nadir position.}
    \label{fig:snapshot}
\end{figure}

\subsection{Satellite Decision and Strategy}
To show the results of heterogeneous agent policy training, we captured HATRPO Policy Execution under Hard Scenario. In the hard scenario, the satellite cluster operates in a highly resource-constrained environment, where limited battery, onboard memory, and communication bandwidth challenge mission success. The HATRPO (Heterogeneous Agent Trust Region Policy Optimization) policy demonstrates coordinated and adaptive decision-making that balances mission objectives (image capturing and downlinking) with resource preservation (see Fig. \ref{fig:snapshot}).

\subsubsection{Coordinated Imaging Strategy}
The satellites employ a distributed yet cooperative imaging pattern, dynamically dividing observation targets among themselves to maximize coverage efficiency. The optical and SAR satellites complement each other. Optical satellites prioritize high-value targets when the AoI's visibility are in a good condition. SAR satellites handle cloudy or night-time regions, ensuring continuity of coverage. The policy avoids redundant imaging by adjusting target assignments in real time, thus conserving both battery energy and memory storage.

\subsubsection{Resource-Aware Decision-Making}

\begin{figure}[t]
    \centering
    \includegraphics[width=\linewidth]{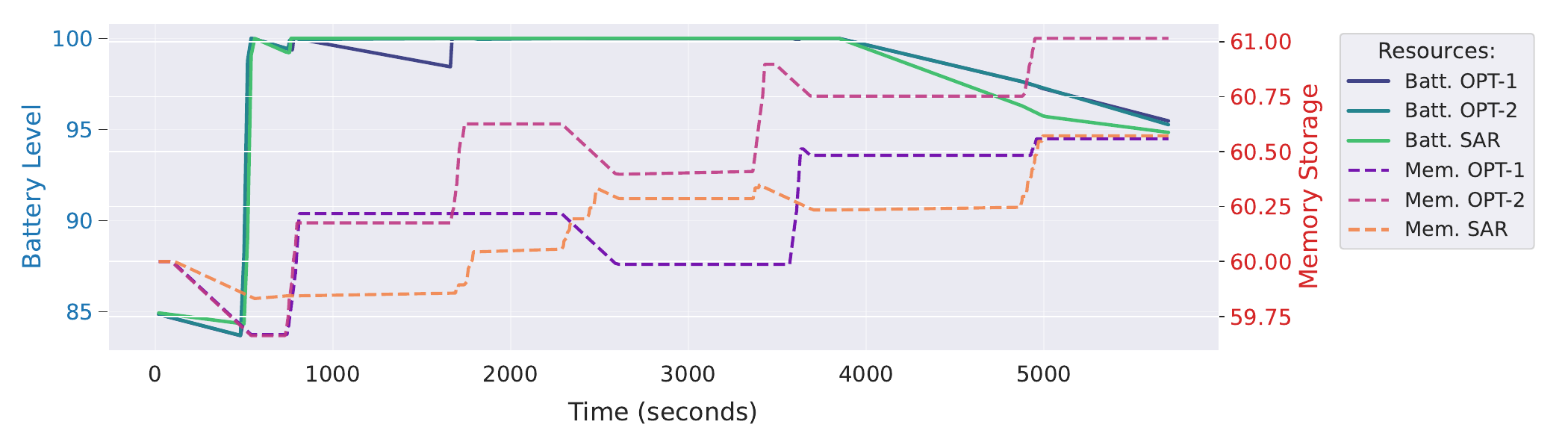}
    \caption{Battery and Memory Resources Utilization of HATRPO Policies in Heterogeneous Cluster}
    \label{fig:hatrpo-res}
\end{figure}

As the result of resource optimization, the HATRPO’s policy explicitly accounts for each satellite’s current battery and memory storage levels (see Fig. \ref{fig:hatrpo-res}). Satellites with lower memory postpone imaging and prioritize downlink tasks when within range of a ground station to offload data efficiently, allowing recovery before resuming operations. The battery level is successfully maintained upper than the minimum allowable level ($>80\%$). While the satellites move under unshaded area, the battery can be charged to increase the battery level and ready to be used for imaging or downlinking. This adaptive behaviour highlights the resource-conditioned policy adaptation, a key strength of HATRPO in maintaining operational stability under harsh conditions.

Overall, the HATRPO-controlled satellite cluster achieves capability to balance trade-off between data acquisition (imaging) and data delivery (downlinking) of the EO mission. It shows resilience against varying resource condition by leveraging heterogeneous policy optimization, allowing each agent (satellite) to act according to its unique capability and current state. Despite the difficulty of the hard scenario, the collective behaviour remains stable and efficient, it illustrates robust cooperative intelligence under heterogeneous resource constraints.

\section{Conclusion}
Our study implement RL and MARL, for heterogenous agents cooperation problem in the context of realistic autonomous satellite cluster resource optimization problem. Through extensive experiments under various scenarios, we demonstrated the effectiveness and adaptability of heterogeneous-agent algorithms (HATRPO and HAPPO) in addressing the challenges of dynamic decision-making and resource optimisation during Earth-observation missions, particularly under restricted-resource conditions. Moreover, the results highlight the potential of heterogeneous MARL in CTDE to enhance collaboration and maintain communication efficiency in satellite operations. In the future, this work can be extended to more complex scenarios, such as larger-scale multi-cluster orbital systems. Additionally, integrating domain-specific knowledge into MARL training and developing methods to further mitigate non-stationarity will be key research directions to enhance real-world deployment feasibility.

\section*{Acknowledgement}
This work has been supported by the SmartSat CRC, whose
activities are funded by the Australian Government’s CRC Program. This work use an open-source realistic satellite simulator (Basilisk and BSK-RL) that is actively developed by Dr. Hanspeter Schaub and team at AVS Laboratory, University of Colorado Boulder. Also, the authors would like to express their sincere gratitude to BAE Systems for their invaluable support and collaboration throughout this research.

\bibliographystyle{elsarticle-harv} 
\bibliography{references}

\end{document}